\documentclass[runningheads]{llncs}
\usepackage[T1]{fontenc}
\usepackage{graphicx}
\usepackage{amsmath}
\usepackage{booktabs}
\usepackage{multirow}
\usepackage{array}
\newcolumntype{C}{>{\centering\arraybackslash}m{3.2em}}

\begin{document}
\title{Graph Property Inference in Small Language Models: Effects of Representation and Reasoning Strategy}
\titlerunning{Graph Property Inference in Small Language Models}
%

\author{
Michal Podstawski\orcidID{0000-0003-1222-6894}
}
    
\authorrunning{M. Podstawski}

\institute{
NASK National Research Institute, Warsaw, Poland\\
\url{https://nask.pl/}
\\
\email{michal.podstawski@nask.pl}
}

\maketitle              
\begin{abstract}

Recent progress in language modeling has expanded the range of tasks that can be approached through natural language interfaces, including problems that require structured reasoning. However, it remains unclear how effectively limited-capacity language models can infer formal properties of relational structures when those structures are presented in textual form.

We conduct a systematic study of graph-theoretic property inference in small instruction-tuned language models, isolating the roles of input representation and reasoning strategy. Across a diverse set of local and global graph metrics evaluated on three models, we find that small language models fail to achieve reliable graph property estimation: normalized errors consistently exceed the intrinsic dispersion of target properties, and rank correlations remain weak across all configurations. However, the failure is structured rather than uniform. Adjacency-list encodings consistently reduce error and improve ordinal consistency relative to edge-lists, and multi-branch reasoning yields measurable aggregate gains across configurations.

These results show that without task-specific fine-tuning or architectural adaptation, graph property inference in pretrained small language models remains fundamentally unreliable, but that representational organization and inference design produce consistent differences. The findings characterize the conditions under which structured inference degrades and identify which design choices yield improvements even under constrained model capacity.


\end{abstract}
\section{Introduction}

Language models are increasingly applied to tasks that require reasoning over relational data. Yet these models operate over linear token sequences rather than explicit structural representations, raising the question of how effectively they can infer formal properties of structured inputs.

Graphs offer a controlled and analytically tractable setting for investigating this question. They provide precisely defined relational structures with well-characterized local and global properties, enabling systematic evaluation across heterogeneous structural metrics. By varying graph topology while keeping representation format explicit, one can directly assess how language models respond to structured variation under controlled conditions.

Although prior work has explored graph-related reasoning tasks, systematic evidence on graph-theoretic property estimation remains limited. In particular, it is unclear how strongly performance depends on the choice of serialization format and the inference strategy used to elicit predictions. Advances such as Chain-of-Thought (CoT)~\cite{ref_cot} and Graph of Thoughts (GoT)~\cite{ref_got} suggest that reasoning structure can materially influence outcomes, but their interaction with representational design in graph settings has not been carefully isolated.

In this work, we conduct a controlled empirical study of graph property inference in small language models (SLMs). We compare adjacency-list and edge-list serializations under direct prediction, CoT reasoning, and GoT aggregation, evaluating performance across heterogeneous graph metrics.

Our results show that without task-specific fine-tuning, all evaluated configurations fail to achieve reliable graph property estimation - normalized errors consistently exceed the intrinsic variability of target properties, and rank correlations remain weak. However, the failure is structured: adjacency-list encodings yield lower normalized error and higher rank consistency than edge-lists, and multi-branch aggregation provides the most consistent improvements. Together, these findings clarify both the limitations of prompting-based graph reasoning in small language models and the design factors that modulate performance within this regime of overall weak estimation.

\section{Related Work}

\subsection{Reasoning in Language Models}

Prompt-based reasoning has significantly advanced the problem-solving capabilities of language models. Chain-of-Thought (CoT) prompting demonstrates that explicitly eliciting intermediate reasoning steps substantially improves performance on arithmetic, symbolic, and logical reasoning tasks \cite{ref_cot,ref_zero_shot_cot}. Subsequent work introduced self-consistency decoding, which aggregates multiple reasoning paths to improve robustness and accuracy \cite{ref_selfconsistency}. 

More recently, structured multi-branch reasoning frameworks such as Graph of Thoughts (GoT) generalize linear reasoning chains into graph-structured reasoning traces, enabling aggregation over alternative inference paths \cite{ref_got}. These approaches show that reasoning structure and aggregation mechanisms can materially influence performance, particularly for tasks requiring distributed evidence integration.

However, most evaluations of these techniques focus solely on mathematical benchmarks or commonsense reasoning tasks \cite{ref_gsm8k,ref_bigbench}. Their effectiveness for graph-theoretic property estimation, especially in small language models, remains largely underexplored.

\subsection{Graph Reasoning with Language Models}

Recent work has investigated the ability of language models to perform graph-related reasoning tasks, including shortest-path computation, connectivity analysis, and symbolic graph manipulation~\cite{ref_graph_survey_llm,ref_llm_combinatorial}. These studies systematically evaluate structural reasoning capabilities and reveal both emergent competence and significant limitations \cite{ref_graph_understanding}. Additional analyses highlight challenges in algorithmic generalization and multi-hop reasoning \cite{ref_algorithmic_reasoning}.

While these works primarily evaluate task-level accuracy on individual reasoning problems, less attention has been given to systematic estimation of heterogeneous graph-theoretic metrics. Moreover, prior studies largely focus on large-scale models, leaving the structural inference capabilities of small-scale SLMs insufficiently characterized.

A notable exception is TinyGraphEstimator~\cite{ref_tinygraph}, which demonstrates that small language models fine-tuned with LoRA achieve strong in-distribution performance on graph property estimation, substantially outperforming zero-shot baselines. Subsequent work~\cite{ref_generalization} extends these findings by showing that fine-tuned SLMs generalize across structurally distinct graph families and to graphs substantially larger than those seen during training, while also identifying a locality gradient in structural understanding. Together, these results establish that fine-tuned small models are competent graph property estimators - raising the question of whether comparable performance can be achieved through prompting and inference strategies alone, without task-specific fine-tuning.

\subsection{Structured Representations and Inductive Bias}

Transformers process token sequences without explicit structural inductive bias \cite{ref_transformer}. Consequently, the serialization format of structured inputs can significantly influence model behavior. Prior work on graph neural networks (GNNs) demonstrates that architectures with explicit relational inductive bias are well suited for structured data \cite{ref_gnn_survey,ref_wl_gnn}. Graph transformer variants further attempt to incorporate structural encodings directly into attention mechanisms \cite{ref_graphormer}.

In contrast, language models must infer structure implicitly from textual encodings. Research in structured prompting and code representation suggests that grouping semantically related elements improves reasoning stability and locality of attention \cite{ref_structured_prompting}. Adjacency-list encodings naturally cluster neighbor information by node, potentially aligning better with transformer attention mechanisms than edge-list representations, which fragment relational information across token pairs.

Our work builds on these insights by empirically isolating the effects of serialization format and inference strategy on graph property estimation in small language models.

\subsection{Positioning of This Work}

Unlike prior studies that primarily evaluate discrete graph reasoning tasks or single-problem accuracy, we perform a controlled multi-factor analysis across:

\begin{itemize}
    \item Representation format (adjacency vs.\ edge-list)
    \item Inference strategy (baseline, CoT, GoT)
    \item Multiple heterogeneous graph properties
\end{itemize}

By combining regression-based error metrics with ordinal consistency measures, our study provides a systematic characterization of structural sensitivity in small language models. This enables a more nuanced understanding of how representation design and reasoning scaffolding affect graph-theoretic inference under limited model capacity.

\section{Experimental Protocol}
\label{sec:protocol}

\paragraph{Design Objective.}
The experimental setup isolates two factors affecting structural inference in small language models: (i) graph serialization format and (ii) inference strategy. Other components - model size, decoding configuration, output constraints, and evaluation metrics - are held constant across conditions.

\paragraph{Models.}
We evaluate three small instruction-tuned transformer models: Llama-3.2-3B-Instruct~\cite{ref_llama32}, Phi-4-mini-instruct~\cite{ref_phi4}, and Qwen2.5-3B-Instruct~\cite{ref_qwen25}. The models were selected as representative high-performing instruction-tuned models in the 3--4B parameter class, based on publicly reported rankings on the LLM Leaderboard~\cite{ref_llm_leaderboard}. Using instruction-tuned models minimizes variance due to formatting instability, allowing clearer isolation of representation and inference effects (the \textit{-Instruct} suffix is omitted hereafter for brevity).

\paragraph{Data.}
Evaluation is conducted on benchmark graphs from the TinyGraphEstimator dataset~\cite{ref_tinygraph}. The dataset consists of undirected, unweighted, and connected graphs spanning a range of sizes and structural configurations. For each graph, we evaluate twelve graph-theoretic properties covering degree statistics, clustering measures, path-based metrics, and discrete invariants. Specifically, these include: \textit{minimum degree, maximum degree, mean degree, degree standard deviation, graph density, average clustering coefficient, global clustering coefficient (transitivity), triangle count, average shortest path length, diameter, chromatic number, and global efficiency}.

\paragraph{Graph Serialization.}
Each graph is serialized in one of two formats:

\begin{itemize}
    \item \textbf{Adjacency-list (Adj):} node-wise neighbor grouping.
    \item \textbf{Edge-list (Edge):} ordered edge pairs without grouping.
\end{itemize}

Both representations include explicit $(n, m)$ headers to control for graph size information.
The two encodings differ only in token organization, enabling evaluation of locality effects independent of content.

\paragraph{Inference Strategies.}
Three inference modes are implemented:

\begin{itemize}
    \item \textbf{Baseline:} direct prediction under deterministic decoding.
    \item \textbf{Chain-of-Thought (CoT):} structured reasoning prompt under deterministic decoding.
    \item \textbf{Graph of Thoughts (GoT):} multi-branch stochastic sampling with aggregation.
\end{itemize}

Baseline and CoT use greedy decoding. GoT generates up to 15 independent branches using temperature $T=0.7$ to ensure sufficient diversity across reasoning paths while maintaining coherent outputs. Branch predictions are aggregated by arithmetic mean to produce the final estimate. All generations are constrained to JSON format using \texttt{lm-format-enforcer}~\cite{noamgat2023lmformatenforcer} to ensure structural validity and eliminate formatting artifacts.

\paragraph{Evaluation Metrics.}

Performance is evaluated using normalized error and rank-based consistency measures.
For each graph property, we compute two normalized root mean squared error variants:
\[
\text{NRMSE}_{\text{range}} =
\frac{\sqrt{\frac{1}{n}\sum_{i=1}^{n}(\hat{y}_i - y_i)^2}}
{y_{\max} - y_{\min}},
\qquad
\text{NRMSE}_{\text{std}} =
\frac{\sqrt{\frac{1}{n}\sum_{i=1}^{n}(\hat{y}_i - y_i)^2}}
{\sigma(y)},
\]
where $\hat{y}_i$ denotes the predicted value, $y_i$ the ground truth, $y_{\max} - y_{\min}$ the range, and $\sigma(y)$ the empirical standard deviation of the target distribution.
NRMSE$_{\text{range}}$ provides a conservative normalization bounded by the observed extremes, while NRMSE$_{\text{std}}$ scales error relative to the intrinsic dispersion of each property.
Together, the two normalizations offer complementary perspectives: NRMSE$_{\text{range}}$ enables comparison across properties with differing scales, while values of NRMSE$_{\text{std}} > 1.0$ indicate that prediction error exceeds the natural variability of the target property.
Macro-level comparisons aggregate both metrics across properties.

To assess ordinal structural sensitivity independently of magnitude accuracy, we additionally report Spearman rank correlation ($\rho$).
This measures whether models preserve the relative ordering of graphs with respect to a given property, even when absolute estimates are imperfect.

\section{Results}

\subsection{Overall Accuracy and Structural Sensitivity}

Table~\ref{tab:macro} summarizes macro-averaged performance across graph properties.
Performance varies across heterogeneous graph metrics, reflecting differing levels of structural difficulty.

\begin{table*}[t]
\centering
\caption{Macro-averaged performance. Lower is better for NRMSE metrics, higher is better for Spearman~$\rho$.}
\label{tab:macro}
\resizebox{\textwidth}{!}{%
\begin{tabular}{ll*{9}{C}}
\toprule
& & \multicolumn{3}{c}{NRMSE\textsubscript{range} $\downarrow$} & \multicolumn{3}{c}{NRMSE\textsubscript{std} $\downarrow$} & \multicolumn{3}{c}{Spearman $\rho$ $\uparrow$} \\
\cmidrule(lr){3-5} \cmidrule(lr){6-8} \cmidrule(lr){9-11}
Model & Rep. & Baseline & CoT & GoT & Baseline & CoT & GoT & Baseline & CoT & GoT \\
\midrule
\multirow{2}{*}{Llama-3.2-3B}
  & Edge & 0.441 & 0.419 & 0.383 & 1.822 & 1.805 & 1.589 & 0.145 & 0.202 & 0.212 \\
  & Adj  & 0.427 & 0.418 & 0.397 & 1.768 & 1.738 & 1.634 & 0.259 & 0.261 & 0.273 \\
\midrule
\multirow{2}{*}{Phi-4-mini}
  & Edge & 0.437 & 0.412 & 0.387 & 2.237 & 1.943 & 1.821 & 0.158 & 0.174 & 0.187 \\
  & Adj  & 0.411 & 0.402 & 0.382 & 2.134 & 2.090 & 1.720 & 0.164 & 0.185 & 0.197 \\
\midrule
\multirow{2}{*}{Qwen2.5-3B}
  & Edge & 0.283 & 0.279 & 0.276 & 1.436 & 1.391 & 1.302 & 0.246 & 0.262 & 0.266 \\
  & Adj  & 0.279 & 0.274 & 0.272 & 1.319 & 1.223 & 1.207 & 0.261 & 0.277 & 0.308 \\
\bottomrule
\end{tabular}%
}
\end{table*}

Normalized errors (NRMSE$_{\text{std}}$) remain above 1.0 in all 18 configurations, indicating that prediction error consistently exceeds the intrinsic variability of the target properties. Even the best-performing configuration (Qwen2.5-3B, Adj, GoT) achieves an NRMSE$_{\text{std}}$ of only 1.207, meaning predictions remain substantially noisier than the underlying property distributions.

Spearman rank correlations are positive across most settings, reaching up to $\rho = 0.308$ (Qwen2.5-3B, Adj, GoT), indicating some preserved ordinal sensitivity to structural variation across graphs. However, correlations of this magnitude reflect only weak agreement - the models capture broad relative differences but lack the consistency required for reliable ranking.

Overall, these results suggest that pretrained small language models exhibit some awareness of structural variation but fall well short of reliable graph property estimation. For comparison, fine-tuned models on the same benchmark achieve Spearman $\rho > 0.9$~\cite{ref_tinygraph,ref_generalization} - highlighting that the limitation lies in the absence of task-specific training rather than in model capacity alone.

\subsection{Effect of Representation}

Representation format has a systematic impact on performance.
Adjacency-list serialization generally reduces macro-level error relative to edge-list encoding (Table~\ref{tab:repr}).

\begin{table*}[t]
\centering
\caption{Adjacency-list advantage over Edge-list (positive values indicate improvement when using Adjacency-list).}
\label{tab:repr}
\resizebox{\textwidth}{!}{%
\begin{tabular}{l*{9}{C}}
\toprule
& \multicolumn{3}{c}{$\Delta$NRMSE\textsubscript{range} $\uparrow$} & \multicolumn{3}{c}{$\Delta$NRMSE\textsubscript{std} $\uparrow$} & \multicolumn{3}{c}{$\Delta$Spearman $\rho$ $\uparrow$} \\
\cmidrule(lr){2-4} \cmidrule(lr){5-7} \cmidrule(lr){8-10}
Model & Baseline & CoT & GoT & Baseline & CoT & GoT & Baseline & CoT & GoT \\
\midrule
Llama-3.2-3B & 0.014 & 0.001 & $-$0.014 & 0.054 & 0.067 & $-$0.045 & 0.114 & 0.059 & 0.061 \\
Phi-4-mini   & 0.026 & 0.010 & 0.005 & 0.103 & $-$0.147 & 0.101 & 0.006 & 0.011 & 0.010 \\
Qwen2.5-3B  & 0.004 & 0.005 & 0.004 & 0.117 & 0.168 & 0.095 & 0.015 & 0.015 & 0.042 \\
\bottomrule
\end{tabular}%
}
\end{table*}

For Qwen2.5-3B, adjacency encoding improves NRMSE$_{\text{std}}$ by 0.117 under baseline prompting and by 0.095 under GoT aggregation.
For Phi-4-mini, baseline improvements reach 0.103.
For Llama-3.2-3B, improvements are smaller and not always consistent - under GoT, edge-list encoding slightly outperforms adjacency-list ($\Delta = -0.045$).

Adjacency encoding also tends to improve rank consistency, with the largest Spearman gain of 0.114 observed for Llama-3.2-3B under baseline prompting.
However, for Phi-4-mini, Spearman differences between representations are negligible (0.006--0.011).

These results suggest that grouping neighbors by node provides a more favorable input structure for transformer attention mechanisms than unordered edge pairs, facilitating local aggregation of structural information. Yet the representation advantage, while consistent in direction, is modest in magnitude - it narrows the gap but does not bring performance close to reliable estimation.

\subsection{Effect of Inference Strategy}
\begin{figure*}[t]
\centering
\includegraphics[width=\textwidth]{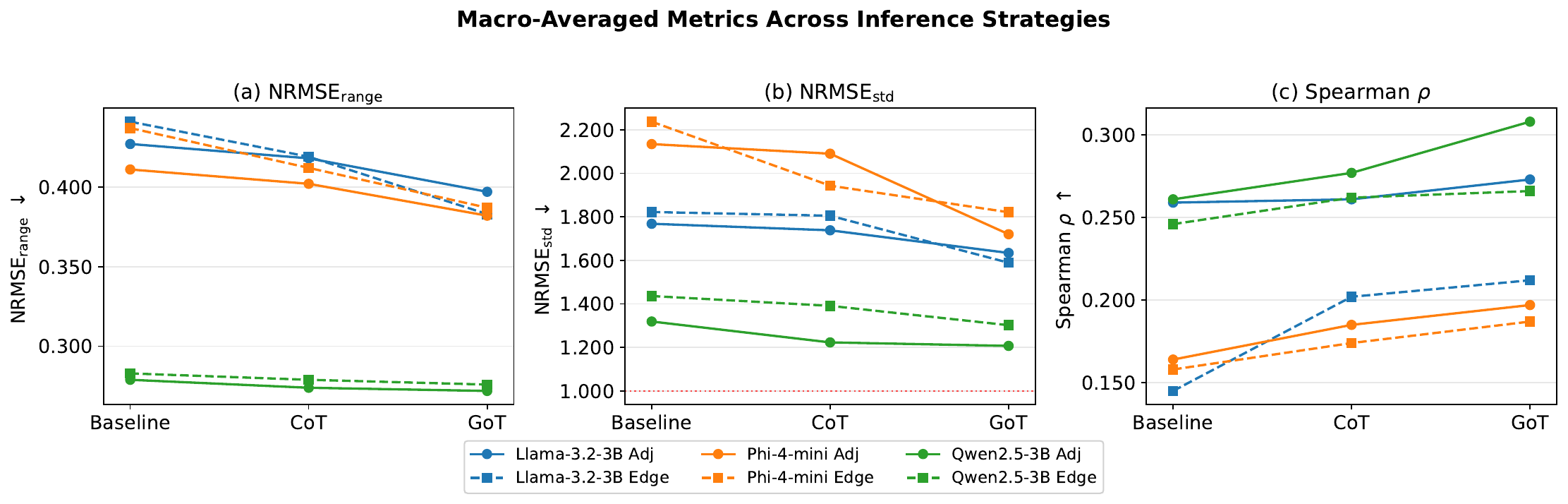}
\caption{Macro-averaged metrics across inference strategies. (a)~Range-normalized error, (b)~standard-deviation-normalized error with the $\text{NRMSE}_{\text{std}} = 1.0$ reference line, and (c)~Spearman rank correlation. Solid lines denote adjacency-list encoding; dashed lines denote edge-list encoding.}
\label{fig:combined}
\end{figure*}

\begin{figure*}[t]
\centering
\includegraphics[width=\textwidth]{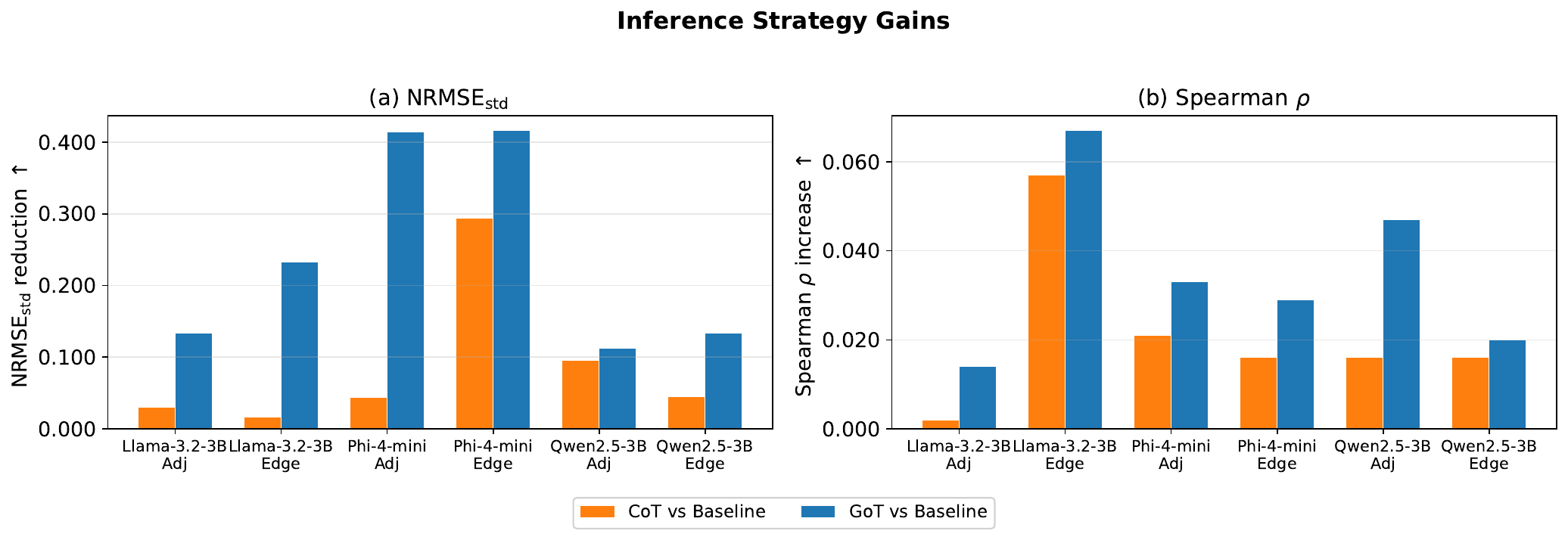}
\caption{Inference strategy improvement relative to baseline prompting. (a)~Reduction in $\text{NRMSE}_{\text{std}}$ (positive values indicate lower error). (b)~Increase in Spearman $\rho$ (positive values indicate improved rank consistency).}
\label{fig:strategy_gains}
\end{figure*}

Inference strategy also influences performance, though the gains remain modest in absolute terms (Table~\ref{tab:strategy}, Figure~\ref{fig:combined}, Figure~\ref{fig:strategy_gains}).
GoT consistently yields the largest improvements relative to baseline prompting.
The largest NRMSE$_{\text{std}}$ reduction is observed for Phi-4-mini under edge-list encoding (0.416), followed by Phi-4-mini under adjacency encoding (0.414).
For Llama-3.2-3B with edge-lists, GoT reduces NRMSE$_{\text{std}}$ by 0.233, while Qwen2.5-3B achieves reductions of 0.134 (Edge) and 0.112 (Adj).
These reductions are accompanied by small improvements in Spearman correlation across most configurations, with the largest gain of 0.067 observed for Llama-3.2-3B with edge-lists.

\begin{table}[t]
\centering
\caption{Inference strategy improvement relative to Baseline. Positive values indicate reduction in NRMSE / increase in Spearman~$\rho$.}
\label{tab:strategy}
\small
\begin{tabular}{ll*{6}{C}}
\toprule
& & \multicolumn{2}{c}{$\Delta$NRMSE\textsubscript{range} $\uparrow$} & \multicolumn{2}{c}{$\Delta$NRMSE\textsubscript{std} $\uparrow$} & \multicolumn{2}{c}{$\Delta$Spearman $\rho$ $\uparrow$} \\
\cmidrule(lr){3-4} \cmidrule(lr){5-6} \cmidrule(lr){7-8}
Model & Rep. & CoT & GoT & CoT & GoT & CoT & GoT \\
\midrule
\multirow{2}{*}{Llama-3.2-3B}
  & Edge & 0.022 & 0.058 & 0.017 & 0.233 & 0.057 & 0.067 \\
  & Adj  & 0.009 & 0.030 & 0.030 & 0.134 & 0.002 & 0.014 \\
\midrule
\multirow{2}{*}{Phi-4-mini}
  & Edge & 0.025 & 0.050 & 0.294 & 0.416 & 0.016 & 0.029 \\
  & Adj  & 0.009 & 0.029 & 0.044 & 0.414 & 0.021 & 0.033 \\
\midrule
\multirow{2}{*}{Qwen2.5-3B}
  & Edge & 0.004 & 0.007 & 0.045 & 0.134 & 0.016 & 0.020 \\
  & Adj  & 0.005 & 0.007 & 0.096 & 0.112 & 0.016 & 0.047 \\
\bottomrule
\end{tabular}
\end{table}

CoT reasoning also improves performance in most configurations, though the gains are smaller and less consistent than those of GoT.
The strongest CoT effect appears for Phi-4-mini with edge-lists ($\Delta\text{NRMSE}_{\text{std}} = 0.294$), while for Llama-3.2-3B the improvements are marginal.
This suggests that while explicit intermediate reasoning steps provide some benefit, multi-branch aggregation is a more effective mechanism for integrating distributed structural information.

Notably, even under the best-performing strategy, all configurations remain well above the NRMSE$_{\text{std}} = 1.0$ threshold. The inference strategy improvements are real and reproducible, but they reduce the severity of estimation failure rather than resolve it.

\subsection{Summary of Findings}

Overall, the results demonstrate three consistent patterns:
\begin{itemize}
    \item Estimation remains unreliable: all configurations produce NRMSE$_{\text{std}}$ above 1.0 and low rank correlations, indicating that pretrained small language models struggle to infer graph-theoretic properties from textual serializations.
    \item Representation matters: adjacency-list encoding consistently reduces error and improves ordinal consistency relative to edge-lists, although overall performance remains limited.
    \item Multi-branch reasoning helps: GoT yields the most consistent improvements across models and metrics, though the absolute gains are modest.
\end{itemize}
Together, these findings indicate that small language models exhibit some sensitivity to structural variation, but that this sensitivity does not translate into reliable inference. The observed improvements from representation and inference design are consistent and reproducible, yet insufficient to overcome the fundamental limitations of prompting-based graph reasoning at this model scale.

\section{Discussion}

The results reveal that pretrained small language models perform poorly on graph-theoretic property inference, but that this poor performance is not monolithic - it is systematically shaped by both input representation and inference strategy.

The most consistent finding is the influence of serialization format. Adjacency-list encoding reduces normalized error and improves rank consistency across all three evaluated models. This suggests that grouping neighbors by node better supports local aggregation of structural information within transformer architectures. Edge-list representations, by contrast, distribute relational information across disjoint token pairs, increasing attention fragmentation. This result is consistent with prior findings on fine-tuned models, where adjacency-list formats similarly yielded more accurate structural reasoning \cite{ref_generalization}, suggesting the format advantage generalizes beyond task-specific training. However, even with adjacency-list encoding, all configurations remain above the NRMSE$_{\text{std}} = 1.0$ threshold, indicating that the representation advantage reduces error without resolving the underlying estimation problem.

Inference strategy further modulates performance. GoT aggregation yields the most consistent and largest improvements across all models, suggesting that its multi-branch structure helps integrate distributed structural cues that are difficult to consolidate through a single reasoning trace. CoT reasoning also improves performance in most configurations, though the gains are smaller and less consistent. This indicates that while explicit intermediate reasoning steps provide some benefit, multi-branch aggregation is a more effective mechanism for structured inference under limited model capacity.

Taken together, these findings suggest that while pretrained small language models are sensitive to structural variation in their inputs, this sensitivity remains shallow. The observed effects of representation and inference design are reproducible and informative for system design, but they operate within a regime of overall weak performance. Bridging the gap to reliable inference likely requires approaches beyond prompting, such as task-specific fine-tuning or architectural adaptation.

\section{Limitations}
This study isolates representation and inference strategy under controlled conditions, but several factors constrain the scope of the conclusions.

First, the evaluation is restricted to three 3--4B-scale instruction-tuned models. Although this design enables focused analysis of limited-capacity systems, the findings may not generalize directly to larger architectures, which may exhibit qualitatively different structural reasoning capabilities.

Second, graphs are serialized as text without explicit structural inductive bias.
Although this reflects realistic language-model usage, it inherently constrains representational efficiency compared to graph neural networks or specialized architectures.

Third, evaluation focuses on property estimation rather than procedural graph reasoning (e.g., explicit shortest-path construction). Future work could examine step-wise structural reasoning to better understand intermediate representation dynamics.

\section{Conclusion}

This study provides a systematic empirical analysis of graph property inference in pretrained small language models, evaluated across three models, two serialization formats, and three inference strategies.

The central finding is negative: without task-specific fine-tuning, small language models fail to achieve reliable graph property estimation. Normalized errors exceed the intrinsic variability of target properties in all 18 evaluated configurations, and rank correlations remain weak throughout.

Within this regime of overall failure, however, the design factors explored produce consistent and reproducible effects. Adjacency-list serialization reduces error and improves ordinal consistency relative to edge-lists across most configurations, suggesting that node-grouped representations better support local structural aggregation in transformer architectures. Chain-of-Thought reasoning helps in most settings, though less consistently than Graph of Thoughts aggregation, which yields the largest improvements across models, indicating that multi-branch reasoning partially compensates for the difficulty of integrating distributed structural cues through a single inference pass.

These improvements are real but modest - they reduce the severity of estimation failure rather than overcome it. The results suggest that prompting and inference strategies alone are insufficient for structured graph reasoning at the 3--4B parameter scale, and that bridging the gap to reliable performance requires approaches such as task-specific fine-tuning or architectural adaptation. More broadly, the findings highlight that structural competence in small language models is shaped not only by model capacity, but by the interaction of representational design, inference strategy, and domain-specific training.

\subsubsection*{Acknowledgements}

This manuscript acknowledges the use of Claude Opus 4.6~\cite{claude} language model developed by Anthropic, to improve language clarity, refine sentence structure, and enhance overall writing precision.

%
%
%
%

\end{document}